\documentclass[runningheads]{llncs}

\usepackage{eccv}

\usepackage{eccvabbrv}
\usepackage[ruled,vlined]{algorithm2e}
\usepackage{graphicx}
\usepackage{booktabs}
\usepackage{multirow} 

\usepackage[accsupp]{axessibility}  

\usepackage[pagebackref,breaklinks,colorlinks,citecolor=eccvblue]{hyperref}

\begin{document}

\title{Ocean4D: Generative Underwater 4D Reconstruction via Medium-Aware Video Diffusion} 

\titlerunning{Ocean4D: Generative Underwater 4D Reconstruction}


\author{Yuqiang Huang\inst{1} \and
	Yuxi Wang\inst{1}\textsuperscript{\dag,\ddag} \and
	Junyu Dong\inst{1} \and
	Zhaoxiang Zhang\inst{2}}

\authorrunning{Y. Huang et al.}

\institute{
	Faculty of Information Science and Engineering, Ocean University of China
	\and
	Institute of Automation, Chinese Academy of Sciences
}
\maketitle
\begingroup
\renewcommand\thefootnote{}
\footnotetext{\textsuperscript{\dag}Corresponding author. \textsuperscript{\ddag}Project lead.\\ Web: https://anonymous.4open.science/w/Ocean4D/\\ Emails: \texttt{yqhuang@stu.ouc.edu.cn}, \texttt{yuxiwang93@gmail.com}, \texttt{dongjunyu@ouc.edu.cn}, \texttt{zhaoxiang.zhang@ia.ac.cn}.}
\endgroup

\begin{abstract}
Underwater 4D reconstruction remains challenging due to the coupling between degraded light transport in participating media and dynamic water variations. Most existing Methods are developed under in-air assumptions and do not explicitly account for underwater absorption and backscatter. Additionally, near-static assumptions make these approaches sensitive to drifting particles and dynamic distractors , leading to unstable geometry and inconsistent cross-view results. 
To address these issues, we propose a generative framework for underwater 4D reconstruction, named \textbf{Ocean4D}, which is built on two complementary components. Specifically, 4D-GCC constructs 4D geometrically consistent conditioning with improved cross-frame coverage, while the Medium-Aware Block performs implicit medium-aware denoising in the latent diffusion process to stabilize underwater appearance under absorption and scattering. Given a monocular video and target cameras, our method generates videos along the target trajectories while preserving global structure and cross-view consistency. Extensive experiments on both dynamic and static underwater benchmarks demonstrate state-of-the-art performance on underwater reconstruction.
  \keywords{Underwater Reconstruction \and Video Generative Model \and Medium-Aware Diffusion Model}
\end{abstract}

\section{Introduction}
\label{sec:introduction}
Underwater reconstruction is an important yet highly challenging problem. In practice, reliable underwater reconstruction supports a wide range of applications, including marine ecological monitoring and environmental protection \cite{de2022high,hopkinson2020automated}, and provides critical perceptual foundations for localization and navigation of autonomous underwater systems \cite{arain2020close,hernandez2016autonomous}. 
Moreover, a particularly compelling scenario involves reconstructing stable and geometrically consistent 4D representations from casually captured monocular videos, which would enable reliable visualization, analysis, and immersive interaction with dynamic underwater environments.

With the advances of NeRF-based representations~\cite{mildenhall2021nerf,barron2021mip,barron2022mip,barron2023zip,chen2021mvsnerf} and 3DGS-based splatting~\cite{kerbl20233d,chen2024mvsplat,fan2024instantsplat,liang2024analytic}, a large number of underwater reconstruction methods like WaterSplatting~\cite{li2025watersplatting} has emerged in recent years~\cite{liu2024aquatic,yang2025seasplat,du2024udr,sethuraman2023waternerf}. However, these methods typically handle only static scenes and cannot address dynamic underwater scenarios. In parallel, the rise of diffusion models~\cite{blattmann2023stable,blattmann2023align} has accelerated Generative 4D Reconstruction~\cite{van2024generative,ren2025gen3c,yu2025trajectorycrafter,bai2025recammaster}. Nevertheless, most diffusion-based approaches are developed for in-air settings and implicitly assume clean imaging with zero medium density~\cite{yariv2021volume}, making them difficult to directly adapt to underwater image formation.

Specifically, observations in underwater environments are jointly governed by distance-dependent attenuation of the direct component and volumetric scattering in the medium. For example, pixel intensities arise from the superposition of surface-reflected radiance and accumulated backscatter along the line of sight, and this process varies significantly with depth, wavelength, and the medium state~\cite{akkaynak2018revised,akkaynak2019sea,hu2023overview}. As a result, underwater 3D/4D reconstruction becomes more challenging due to the medium effect. 
Moreover, existing monocular methods for 4D reconstruction typically rely on time-varying re-rendering to construct conditioning signals for novel views. When occlusion or large viewpoint shifts introduce large unseen regions, the conditioning becomes sparse and incomplete. This eventually causes the reconstruction to deviate from the source-video geometry and even induce semantically inconsistent hallucinations in missing regions.

\begin{figure*}[!t]
	\centering
	\includegraphics[width=\linewidth]{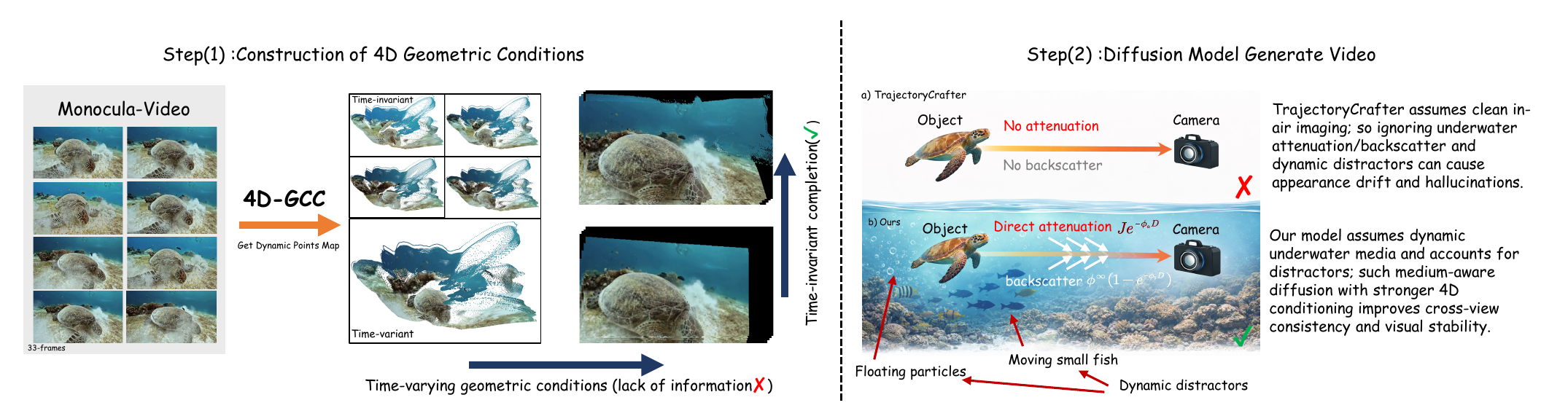}
	\caption{\textbf{Ocean4D} addresses the problem of underwater 4D reconstruction from a monocular video. Ocean4D constructs 4D geometrically consistent conditions and performs medium-aware generation with a video diffusion model, improving cross-view consistency and visual stability under dynamic underwater media.}
	\label{fig:shoutu2}
    \vspace{-0.4cm}
\end{figure*}

To address the above issues, we propose \textbf{Ocean4D} (Fig.~\ref{fig:shoutu2}), a generative framework for underwater 4D reconstruction from monocular videos. Ocean4D is built upon two complementary components. First, \textbf{4D-GCC} (4D Geometrically Consistent Conditioning) provides a geometry-centric conditioning abstraction that aggregates cross-frame geometric evidence into a consistent 4D representation, improving conditioning completeness under occlusions and large viewpoint shifts. Second, we introduce a \textbf{Medium-Aware Block} that performs \emph{implicit} medium modeling inside the diffusion denoising process. Instead of explicit medium-parameter regression, the Medium-Aware Block introduces depth-aware modulation into the denoising process, making the generator more robust to absorption-induced and scattering-induced appearance variations. Together, 4D-GCC and the Medium-Aware Block enable Ocean4D to produce geometry-consistent and medium-stable 4D reconstructions in challenging underwater scenarios. 
We conduct extensive evaluations on both dynamic and static underwater benchmarks, and the results demonstrate that Ocean4D achieves state-of-the-art performance in terms of geometric consistency and underwater visual quality. Our main contributions are summarized as follows:
\begin{itemize}
	\item We introduce \textbf{4D-GCC} for 4D geometrically consistent conditioning, improving conditioning completeness under occlusions and large viewpoint shifts.
	\item We propose a \textbf{Medium-Aware Block} for implicit medium-aware denoising, stabilizing underwater appearance under absorption, scattering, and medium variation.
	\item We perform comprehensive experiments on dynamic and static underwater benchmarks, demonstrating state-of-the-art results on both quantitative and qualitative metrics.
\end{itemize}

\section{Related Work}
\label{sec:related_work}

\noindent\textbf{3D/4D Scene Reconstruction.} 
\label{sec:related_recon}
3D reconstruction and novel view synthesis have progressed rapidly with both implicit and explicit scene representations. NeRF-based methods recover continuous radiance fields via volume rendering and have been advanced with anti-aliasing and multi-scale designs (e.g., the Mip-NeRF series)~\cite{barron2021mip,barron2022mip,barron2023zip}, as well as multi-view and generalized extensions~\cite{chen2021mvsnerf,feng2025ae,verbin2024ref,yu2023nofa}. In parallel, 3D Gaussian Splatting (3DGS) \cite{kerbl20233d} represents scenes with explicit Gaussian primitives and differentiable splatting, enabling efficient rendering and editable representations; follow-up works further improve reconstruction quality through better initialization, geometric constraints, and rendering formulations~\cite{chen2024mvsplat,fan2024instantsplat,liang2024analytic,zhu2024fsgs,lin2021barf}. Dynamic and 4D settings typically require modeling time-varying geometry and appearance, and are commonly tackled via deformation or spatiotemporal representations~\cite{pumarola2021d,park2021nerfies,song2023nerfplayer,duan20244d,wu20244d}.

Meanwhile, feed-forward large reconstruction models (LRMs) and COLMAP-free paradigms have substantially improved geometry recovery from monocular or sparse inputs by leveraging transformer representations and cross-view matching~\cite{wang2024dust3r,zhang2024monst3r,wang20253d,wang2025continuous}. Building on these priors, AnySplat~\cite{jiang2025anysplat} combines VGGT~\cite{wang2025vggt} with splatting-based representations to enable fast reconstruction without per-scene optimization, providing stronger geometric cues that can be used as conditioning under complex degradations.

\noindent\textbf{Underwater Scene Reconstruction.} 
\label{sec:related_underwater}
Early studies often focus on image enhancement and color correction, aiming to recover more natural visual appearances without using explicit models, e.g., methods based on contrast/color distribution adjustment or Retinex-style strategies~\cite{iqbal2010enhancing,ghani2014underwater,fu2014retinex,zhang2017underwater}. More recently, an increasing number of works incorporate underwater image formation into neural rendering. Represented by SeaThru-NeRF \cite{levy2023seathru} and NURS \cite{tang2024neural} approaches, NeRF-based methods explicitly model light transport in participating media by introducing medium-related fields or components, improving the physical consistency of underwater novel view synthesis~\cite{sethuraman2023waternerf,chen2024sp}. In parallel, 3DGS-based methods \cite{li2025watersplatting,liu2024aquatic,yang2025seasplat,du2024udr} have also been extended to underwater settings by incorporating medium modeling or learnable medium features into Gaussian primitive rendering, achieving a better trade-off between efficiency and quality. 
Despite substantial progress, most existing underwater reconstruction methods still focus on static scenes and are sensitive to real-world underwater perturbations, which can lead to unstable reconstructions.

\noindent\textbf{Generative Novel View Synthesis.} 
\label{sec:related_generative_nvs}
With the rapid progress of diffusion models for image and video generation~\cite{song2020denoising,blattmann2023stable,chen2023videocrafter1,lin2024open,yang2024cogvideox}, diffusion-based generative novel view synthesis has become an important direction. Representative works typically control camera motion via geometry-related signals, such as conditioning on depth warping or point cloud renderings~\cite{van2024generative,yu2024viewcrafter}, and further develop various conditioning designs~\cite{gao2024cat3d,muller2024multidiff,sargent2024zeronvs,wu2024reconfusion}. In contrast to explicit geometric conditioning, Zero-1-to-3~\cite{liu2023zero} adopts pose embeddings for implicit pose control, enabling controllable viewpoint generation without explicitly geometry.

Extending these generative methods to dynamic scenes requires maintaining 4D consistency under both camera motion and scene motion. CAT4D~\cite{wu2025cat4d} constructs 4D scenes by fine-tuning CAT3D~\cite{gao2024cat3d} on monocular videos. Similar to GCD~\cite{van2024generative}, Gen3C~\cite{ren2025gen3c} and TrajectoryCrafter~\cite{yu2025trajectorycrafter} use depth warping as conditioning to guide novel trajectory synthesis. In contrast, ReCamMaster~\cite{bai2025recammaster} directly generates videos given the input video and the target trajectory. While it benefits from strong pretrained diffusion priors, it often lacks geometric constraints under large viewpoint changes, making it difficult to guarantee cross-view consistency.

\section{Method}
\label{sec:method}

\subsection{Preliminaries}
\label{sec:Preliminaries}

\noindent\textbf{Underwater Image Formation.}
\label{sec:image_formation_underwater}
In underwater environments, image formation is jointly affected by distance-dependent energy attenuation and volumetric scattering processes. The direct light reflected from objects undergoes exponential decay during propagation due to absorption and scattering, while volumetric scattering in the medium introduces an additional backscatter component along the line-of-sight. Following the classical underwater image formation model~\cite{akkaynak2018revised}, the observed linear image intensity at each pixel can be expressed as:
\begin{equation}
	I = J \cdot e^{-\phi_a D} + \phi^\infty \cdot \left(1 - e^{-\phi_b D}\right),
	\label{eq:underwater_imaging}
\end{equation}
where $J$ denotes the clear scene radiance in the absence of the medium, $D$ represents the scene depth corresponding to each pixel, $\phi_a$ and $\phi_b$ denote the attenuation coefficients for the direct signal and the backscatter component, respectively, and $\phi^\infty$ represents the scattering color of the water at infinity. The first term in Eq.~\eqref{eq:underwater_imaging} characterizes the exponential attenuation of the object-reflected light with propagation distance. The second term describes the accumulated backscatter contribution along the line of sight, whose intensity increases monotonically with depth and asymptotically approaches $\phi^\infty$ at large distances.

\noindent \textbf{Video Diffusion Models.}
\label{sec:4d_diffusion}
Video diffusion models~\cite{blattmann2023stable,blattmann2023align,chen2023videocrafter1,yang2024cogvideox} aim to model the spatiotemporal distribution of multi-frame video data. Given a video sample
$x_0 \in \mathbb{R}^{F \times 3 \times H \times W}$,
the forward diffusion process constructs a Markov chain by gradually injecting Gaussian noise into the data, smoothly transforming the real data distribution into a Gaussian distribution. This process can be written as:
\begin{equation}
	q(x_t \mid x_{t-1})
	=
	\mathcal{N}
	\left(
	x_t;
	\sqrt{\alpha_t} x_{t-1},
	(1-\alpha_t)\mathbf{I}
	\right),
	\label{eq:4d_forward}
\end{equation}
where $\{\alpha_t\}_{t=1}^T$ is a predefined noise schedule. This leads to the closed-form expression at an arbitrary timestep
\begin{equation}
	x_t
	=
	\sqrt{\bar{\alpha}_t} x_0
	+
	\sqrt{1-\bar{\alpha}_t}\epsilon,
	\quad
	\epsilon \sim \mathcal{N}(0,\mathbf{I}).
	\label{eq:4d_closed}
\end{equation}

The reverse process progressively removes noise by learning a parameterized distribution $p_\theta(x_{t-1} \mid x_t)$. In practice, the noise prediction formulation is commonly adopted, where a neural network $\epsilon_\theta(x_t, t)$ is trained to minimize:
\begin{equation}
	\min_{\theta}
	\;
	\mathbb{E}_{x_0,
		t \sim \mathcal{U}(1,T),
		\epsilon \sim \mathcal{N}(0,\mathbf{I})}
	\left[
	\left\|
	\epsilon_\theta(x_t, t)
	-
	\epsilon
	\right\|_2^2
	\right].
	\label{eq:4d_objective}
\end{equation}
A pretrained 3D VAE~\cite{yang2024cogvideox} is typically used to compress videos into the latent space $z = \mathcal{E}(x)$, and the diffusion process is performed in the latent space to reduce computational complexity. The latent variables are first partitioned into patch tokens and then fed into a DiT~\cite{yu2025trajectorycrafter}. During inference, iterative denoising produces latent variables $\hat{z}$, which are finally decoded into video frames $\hat{x} = \mathcal{D}(\hat{z})$.

\begin{figure*}[!t]
	\centering
	\includegraphics[width=\linewidth]{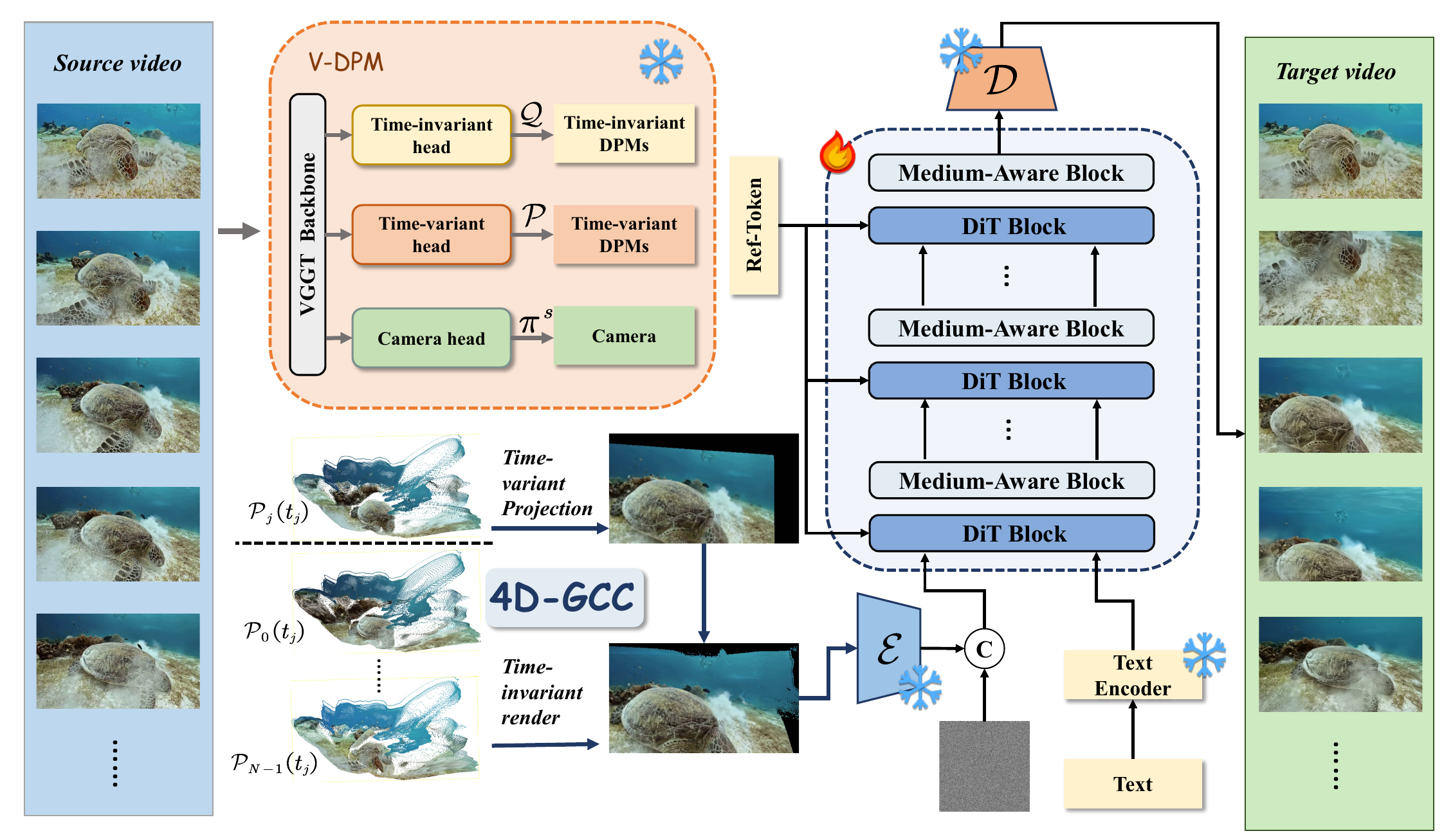}
	\caption{\textbf{Overview of Ocean4D.} Given a monocular source video with camera poses, our method generates multi-view videos under target poses using a video diffusion model. We first apply \textbf{4D-GCC} to extract a 4D geometric representation and construct a geometry-consistent conditioning video by combining time-varying and time-invariant renderings to enrich cross-frame information. The resulting conditioning tokens, together with reference tokens and text tokens from a frozen text encoder, are fed into a diffusion backbone where DiT Blocks and \textbf{Medium-Aware Blocks} are interleaved (DiT Block includes Ref-DiT Block~\cite{yu2025trajectorycrafter}). Finally, the denoised latents are decoded to produce a high-quality target video aligned with the target camera poses and temporally consistent with the source video.}
	\label{fig:ulmd_pipeline}
    \vspace{-0.4cm}
\end{figure*}

\subsection{4D-GCC: 4D Geometrically Consistent Conditioning}
\label{sec:vdpm_condition}


Given a monocular source video $I^s=\{I^s_i\}_{i=0}^{N-1}$ and the corresponding camera pose sequence $\{\pi^s_i\}_{i=0}^{N-1}$, our goal is to generate multi-view results with consistent geometry and temporal dynamics under a specified target camera pose sequence $\{\pi^t_i\}_{i=0}^{N-1}$. To achieve this, the key lies in constructing a complete and geometrically consistent conditioning signal. Previous approaches~\cite{van2024generative,yu2025trajectorycrafter} typically rely on depth priors~\cite{hu2025depthcrafter} to render geometry from source frame $i$, causing the condition for target frame $i$ to depend primarily on source frame $i$. Under occlusions, viewpoint changes, and rapid motion, this results in large invisible regions, leading to sparse and unstable conditioning information. To address this issue, we introduce \textbf{4D-GCC} (4D Geometrically Consistent Conditioning), which extracts a 4D geometric representation from the input video and explicitly enhances cross-frame geometric complementarity, thereby providing more complete conditioning signals for target views (Fig.~\ref{fig:ulmd_pipeline}).

Let $\pi_0$ denote a fixed reference viewpoint (typically the viewpoint of the first source frame). 4D-GCC takes $N$ input frames and predicts two sets of point maps: a time-varying point map $\mathcal{P}$, representing 3D points at each frame's own timestamp; and a time-invariant point map $\mathcal{Q}$, aligning all frames to a common reference timestamp $t_j$. This process can be expressed as:
\[
(\mathcal{P}, \mathcal{Q}) = \Psi(I_0^s, \ldots, I_{N-1}^s).
\]
Here, $\Psi$ denotes the point-map predictor used in 4D-GCC. In our implementation, we instantiate $\Psi$ with V-DPM~\cite{sucar2026v}. $\mathcal{P} = \{\mathcal{P}_i(t_i, \pi_0)\}_{i=0}^{N-1}$ denotes the time-varying point clouds, and $\mathcal{Q}_j = \{\mathcal{P}_i(t_j, \pi_0)\}_{i=0}^{N-1}$ denotes the time-invariant point clouds, where $\mathcal{P}_i(t, \pi) \in \mathbb{R}^{3 \times H \times W}$. Specifically, $\mathcal{P}_i(t_i, \pi_0)$ represents the 3D point cloud of frame $i$ at timestamp $t_i$ under the reference viewpoint $\pi_0$. Given camera intrinsics $K \in \mathbb{R}^{3 \times 3}$, we can transform the 3D points from the reference viewpoint $\pi_0$ to the target viewpoint $\pi^t_i$, which can be written as:
\begin{equation}
	\tilde{I}_i^v = \Phi\left( \pi^t_i \cdot \mathcal{P}_i(t_i, \pi_0); K \right), \quad
	\tilde{I}_{j,i}^q = \Phi\left( \pi^t_i \cdot \mathcal{P}_j(t_i, \pi_0); K \right),
	\label{eq:render_both}
\end{equation}
where $\Phi$ denotes the projection operation. Through Eq.~\eqref{eq:render_both}, we obtain the primary renderings based on time-varying point maps,
$\tilde{I}^v=\{\tilde{I}_i^v\}_{i=0}^{N-1}$,
as well as auxiliary renderings generated from cross-frame aligned point maps,
$\tilde{I}^q=\{\tilde{I}_{j,i}^q\}_{i,j=0}^{N-1}$.
For target view $i$, the time-varying rendering $\tilde{I}_i^v$ provides direct geometric information from source frame $i$, but under large viewpoint shifts or rapid motion, its coverage is often limited. In contrast, the time-invariant rendering $\tilde{I}_{j,i}^q$, explicitly aligned to timestamp $t_i$, introduces geometry from other source frames $j\neq i$ that may be invisible in the current view but observable across frames. Let $\tilde{M}_i^v$ and $\tilde{M}_{j,i}^q$ denote the corresponding visibility masks. The target view is then obtained as:
\begin{equation}
	\hat{I}_i
	=
	\tilde{I}_i^v \big|_{\tilde{M}_i^v}
	+
	\sum_{j\neq i}
	\tilde{I}_{j,i}^q \big|_{\tilde{M}_{j,i}^q \cap \overline{\tilde{M}_i^v}}.
	\label{eq:target_fusion}
\end{equation}

After obtaining the fused target view sequence
$\hat{I}=\{\hat{I}_i\}_{i=0}^{N-1}$,
we map it into the conditional latent space of the diffusion model. Let $\mathcal{E}_{\mathrm{VAE}}(\cdot)$ denote a pretrained VAE encoder. The corresponding conditional latent variables are:
\begin{equation}
	z_i^{\mathrm{cond}}
	=
	\mathcal{E}_{\mathrm{VAE}}(\hat{I}_i),
	\label{eq:condition_latent}
\end{equation}
which are subsequently used as geometric conditioning inputs to the Video diffusion generation module.

\begin{figure*}[t]
	\centering
	\includegraphics[width=\linewidth]{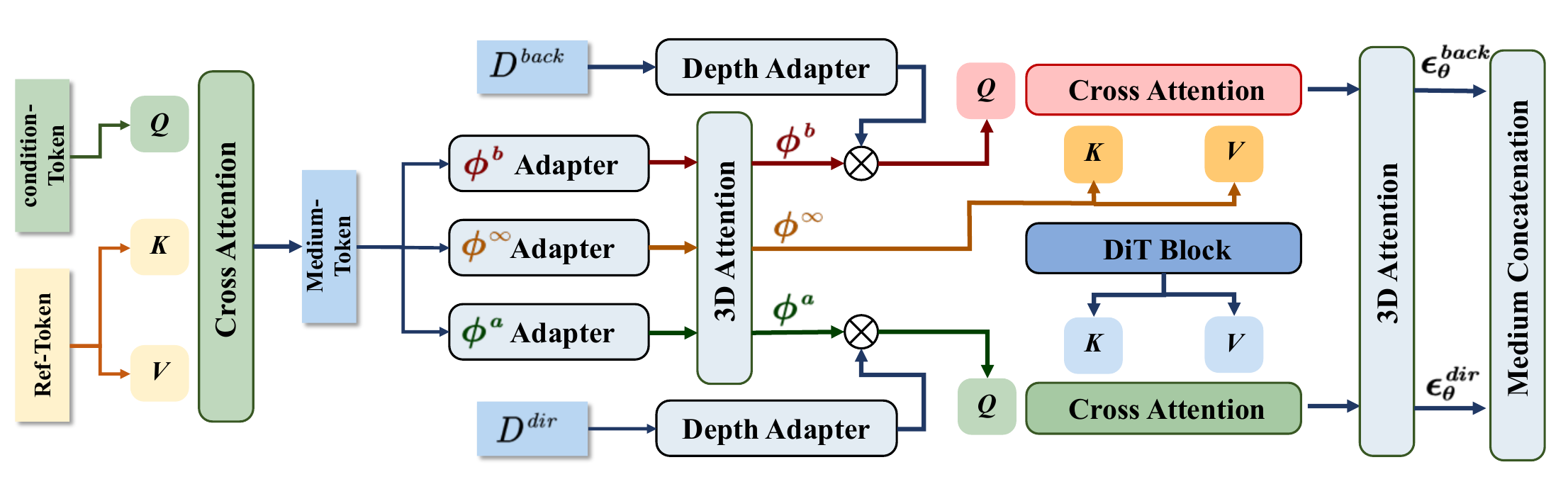}
	\caption{\textbf{Medium-Aware Block architecture.} The Medium-Aware Block takes the conditioning tokens and reference tokens as inputs and first produces a medium token via cross-attention. Three lightweight adaptors map this token to medium-related parameters $\{\phi^a,\phi^b,\phi^\infty\}$. Depth-dependent components $D^{\mathrm{dir}}$ and $D^{\mathrm{back}}$ are injected through depth adaptors to modulate the direct and backscatter pathways. Two cross-attention branches predict $\epsilon_\theta^{\mathrm{dir}}$ and $\epsilon_\theta^{\mathrm{back}}$, which are then concatenated to form the final medium representation for denoising.}
	\label{fig:medium_block}
    \vspace{-0.4cm}
\end{figure*}

\subsection{Implicit Medium Modeling in Latent Diffusion} 
\label{sec:medium_diffusion}

In Sec.~\ref{sec:vdpm_condition}, we have obtained the fused view sequence $\hat{I}=\{\hat{I}_i\}_{i=0}^{N-1}$ aligned with the target camera trajectory $\{\pi_i^t\}_{i=0}^{N-1}$, along with its corresponding geometric conditioning latent variables $z_i^{\mathrm{cond}}=\mathcal{E}_{\mathrm{VAE}}(\hat{I}_i)$. By projecting the temporally aligned point clouds $\mathcal{P}$, we can acquire depth values $D=\{D_i\}_{i=0}^{N-1}$. We now proceed to learn the conditional distribution of the target video $I^t$ in the latent space. Specifically, we leverage CogVideoX's VAE~\cite{yang2024cogvideox} for video compression and decoding, use BLIP~\cite{li2022blip} and T5~\cite{raffel2020exploring} for extracting text-based semantic embeddings, and utilize the DiT backbone from TrajectoryCrafter~\cite{yu2025trajectorycrafter} for spatiotemporal diffusion.

Based on the underwater imaging formula in Eq.~\eqref{eq:underwater_imaging}, the image is composed of direct and backscatter components. The parameters $\phi_a$, $\phi_b$, and $\phi^\infty$ represent the backscatter coefficient, forward scattering coefficient, and background light, respectively, with each parameter corresponding to a value for each of the three color channels, totaling nine parameters. One straightforward approach is to predict $\{\phi_a,\phi_b,\phi^\infty\}$ using a small head from the source video and explicitly model the direct and scatter components in the output image. However, in dynamic water bodies and across-view generation scenes, the medium parameters change over time and space, and explicit regression fails to stabilize.

Therefore, we propose a \textbf{Medium-Aware Block} to perform implicit medium modeling in the latent space.We treat the target video latent $z_0=\mathcal{E}_{\mathrm{VAE}}(I^t)$ as being jointly shaped by a direct component and a backscatter component, and embed this decomposition into the diffusion denoising process. Hence, we formulate the noise prediction as:
\begin{equation}
	\epsilon_\theta(z_t,t;\mathcal{C})
	= 
	\mathcal{M}\!\left(
	\epsilon_\theta^{\mathrm{dir}}(z_t,t;\mathcal{C}),
	\epsilon_\theta^{\mathrm{back}}(z_t,t;\mathcal{C})
	\right),
	\label{eq:implicit_medium}
\end{equation}
where $\mathcal{C}$ represents the conditioning set of the diffusion model, including geometric conditioning latent variables $z_i^{\mathrm{cond}}$, the depth sequence $D_i$, camera pose information, and other relevant conditions. $\epsilon_\theta^{\mathrm{dir}}$ corresponds to the direct denoising component, and $\epsilon_\theta^{\mathrm{back}}$ corresponds to the backscatter component. This formulation does not explicitly correspond to Eq.~\eqref{eq:underwater_imaging}, but instead implements structural equivalence at the latent space level, enabling the diffusion network to learn the modulation dynamics of the medium end-to-end.

To realize this implicit modeling, we insert a \textbf{Medium-Aware Block} after each DiT block, as shown in Fig.~\ref{fig:medium_block}. Given the output feature $h_t^{(l)}$ of the DiT, the Medium-Aware Block takes it as input, generating a medium token via the conditioning token and reference token. This token is processed by an adapter to build the parameters $\{\phi_a, \phi_b, \phi^\infty\}$. Since relying solely on image information does not adequately capture the underwater imaging process, we retain the structural form from Eq.~\eqref{eq:underwater_imaging} by introducing depth-related exponential components in the latent space, where $D_{i}^{\mathrm{dir}} = e^{-D_i}$ represents the depth attenuation for the direct component, and $D_{i}^{\mathrm{back}} = 1 - e^{-D_i}$ captures the depth-enhanced trend for the backscatter component. These components are not treated as explicit physical constraints but are incorporated within the Medium-Aware Block, where they are fused with $h_t^{(l)}$ to yield the overall $\epsilon_\theta(z_t,t;\mathcal{C})$. The implicit medium dynamic modeling runs throughout the entire denoising process, not concentrated at the final output stage, and does so without additional physical parameter supervision. The direct and backscatter decompositions are learned adaptively in the latent space by the diffusion network.

The 4D-GCC conditioning introduced in Sec.~\ref{sec:vdpm_condition} provides geometrically consistent structural constraints across views, while the Medium-Aware Block models underwater medium effects implicitly within the diffusion denoising process. Both components work collaboratively in a unified latent space framework, enabling stable 4D reconstruction for underwater scenes.

\section{Experiments}
\label{sec:experiments}

\subsection{Experimental Settings}
\label{sec:exp_settings}

\noindent\textbf{Datasets.}
We use multiple datasets for training and evaluation to cover both static and dynamic underwater scenes, and to mitigate the limited availability of synchronized underwater multi-view data. Our training data come from UVEB~\cite{xie2024uveb} and ReCamMaster~\cite{bai2025recammaster}. UVEB~\cite{xie2024uveb} provides real underwater video samples with diverse degradation distributions, while ReCamMaster~\cite{bai2025recammaster} offers paired supervision across camera types and trajectories. We emphasize that ReCamMaster~\cite{bai2025recammaster} is not an underwater dataset. We incorporate it because high-quality synchronized underwater multi-view data are scarce, and the stable geometry--trajectory correspondences in ReCamMaster~\cite{bai2025recammaster} help constrain 4D spatiotemporal consistency, thereby ensuring correct 4D modeling. \textit{During training}, we use the static-scene subset of UVEB~\cite{xie2024uveb}. This subset contains no prominent dynamic subjects, but allows small-scale motion disturbances such as small fish and bubbles. UVEB~\cite{xie2024uveb} is split by extracting a continuous segment from each static video and evenly dividing it into source and target clips to form training pairs. We also train with ReCamMaster~\cite{bai2025recammaster}, where two camera settings are selected from a predefined set of ten camera types, with one used as the source and the other as the target.

\textit{For evaluation on dynamic scenes}, we select the remaining dynamic subset of UVEB~\cite{xie2024uveb} and keep samples with moderate motion to enable stable comparisons. The static UVEB data used for training are not included in the evaluation. \textit{For evaluation on static scenes}, we use two benchmarks: NUSR~\cite{tang2024neural} and DRUVA~\cite{varghese2023self}. NUSR contains four real underwater scenes and involves more complex distractors, such as fast motions of small fish. DRUVA~\cite{varghese2023self} is captured under natural illumination and includes disturbances introduced by human interference and water flow.

\noindent\textbf{Implementation Details.}
We train our model on 8 \textbf{Nvidia 4090} GPUs. The learning rate is set to $1\times10^{-5}$ for the DiT Block and $4\times10^{-5}$ for the Medium-Aware Block. To ensure fair comparison with competing methods, we standardize the input clip length to 33 frames and resize each frame to a resolution of $480\times832$. For all videos, we estimate camera poses using the camera head provided by V-DPM~\cite{sucar2026v}. During inference, we set both input and output lengths at 33 frames.

\noindent\textbf{Baselines.}
We compare our method with recent representative approaches, including 3D Gaussians \cite{kerbl20233d}, Water-Splatting \cite{li2025watersplatting}, AnySplat \cite{jiang2025anysplat}, TrajectoryCrafter \cite{yu2025trajectorycrafter}, and ReCamMaster \cite{bai2025recammaster}. \textbf{3DGS}\cite{kerbl20233d} is a canonical explicit scene representation with differentiable rendering based on optimizing a set of 3D Gaussian primitives. \textbf{Water-Splatting}\cite{li2025watersplatting} is an important underwater variant of 3DGS that incorporates underwater image formation modeling to jointly optimize scene appearance and medium effects, achieving SOTA performance on static underwater reconstruction. \textbf{AnySplat}\cite{jiang2025anysplat} represents recent feed-forward reconstruction methods. It combines fast geometric priors such as VGGT \cite{wang2025vggt} and predicts a directly renderable Gaussian scene representation in a single forward pass, enabling robust novel view synthesis without per-scene optimization.
In addition, we compare against two diffusion-based generative novel view synthesis methods, \textbf{TrajectoryCrafter}\cite{yu2025trajectorycrafter} and \textbf{ReCamMaster}\cite{bai2025recammaster}. Both leverage video diffusion models for novel-trajectory generation, but differ in how pose conditioning is injected. Among them, TrajectoryCrafter currently achieves SOTA performance on novel-trajectory generation and 4D-consistent modeling for dynamic scenes.

\subsection{Evaluation on Dynamic Underwater Scenes}
\label{sec:exp_dynamic}

\begin{figure*}[t]
	\centering
	\includegraphics[width=0.95\linewidth]{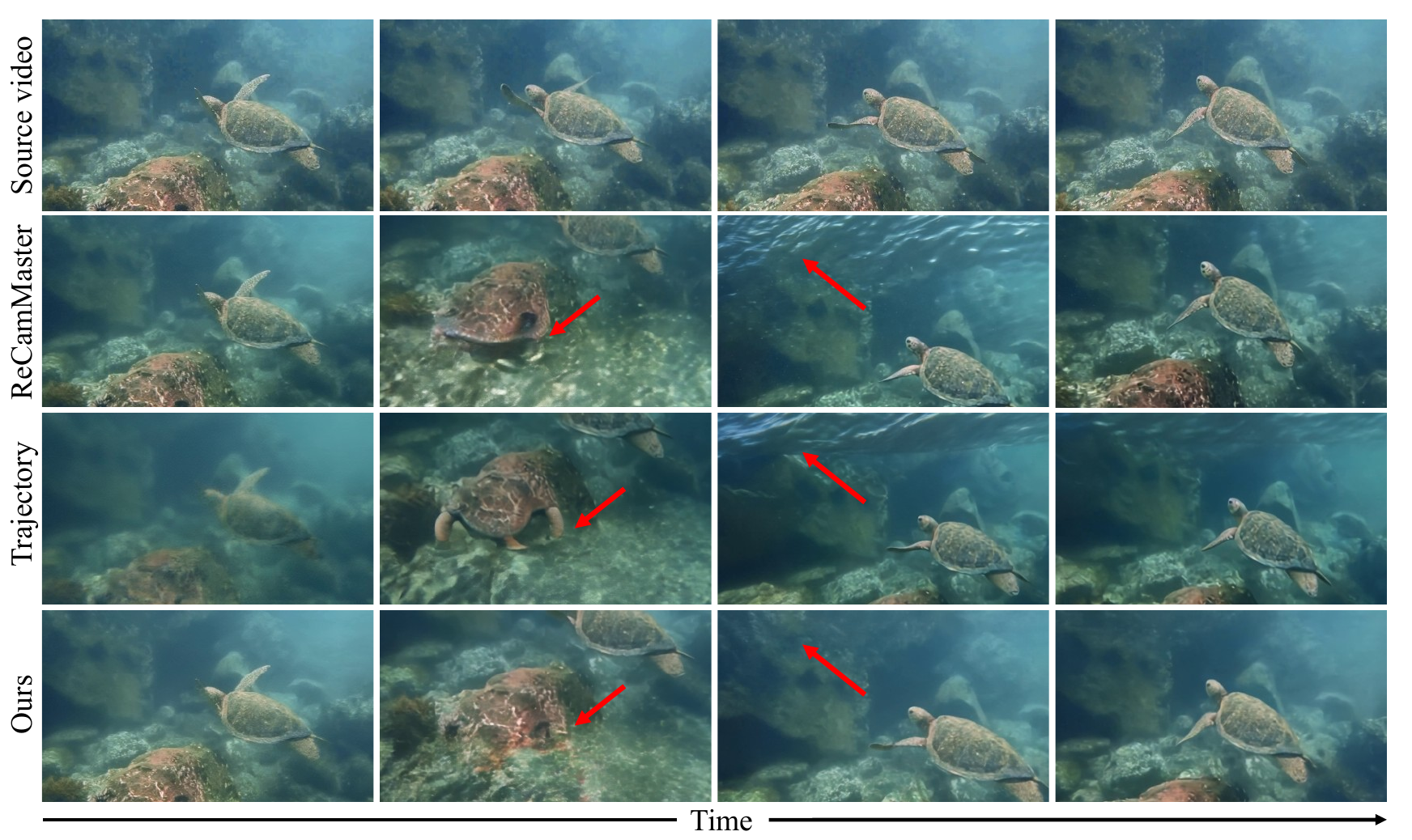}
	\caption{Qualitative results on dynamic underwater scenes comparing our method with TrajectoryCrafter\cite{yu2025trajectorycrafter} and ReCamMaster\cite{bai2025recammaster}. Red arrows highlight typical failure modes under novel views, including semantic misrecognition and hallucinations caused by insufficient geometry.}
	\label{fig:dynamic_compare}
    \vspace{-0.2cm}
\end{figure*}

\begin{table}[t]
	\centering
    \scriptsize
	\caption{{Quantitative comparison with state-of-the-art methods on VBench metrics.}~\cite{huang2024vbench} 
    Best results are in \textbf{bold}, and the second best are \underline{underlined}.}
	\label{tab:vbench_dynamic}
	\setlength{\tabcolsep}{4pt}
	\begin{tabular}{lcccccc}
		\toprule
		\multirow{2}{*}{Method} & Subject & Background  & Temporal & Motion & Aesthetic & Imaging  \\
        & Consistency & Consistency &  Flickering & Smoothness &  Quality & Quality \\
		\midrule
		
		TrajectoryCrafter\cite{yu2025trajectorycrafter} & 0.844 & 0.904 & 0.932 & 0.954 & \underline{0.454} & 0.487 \\
		ReCamMaster\cite{bai2025recammaster} & \underline{0.848} & \underline{0.914} & \underline{0.937} & \textbf{0.989} & 0.436 & \underline{0.528} \\
        \textbf{Ours} & \textbf{0.906} & \textbf{0.920} & \textbf{0.982} & \underline{0.967} & \textbf{0.486} & \textbf{0.591} \\
		\bottomrule
	\end{tabular}
    \vspace{-0.4cm}
\end{table}

\noindent\textbf{Qualitative Comparison.}
As shown in Fig.~\ref{fig:dynamic_compare}, we compare our method with TrajectoryCrafter\cite{yu2025trajectorycrafter} and ReCamMaster\cite{bai2025recammaster} on dynamic underwater scenes from UVEB~\cite{xie2024uveb}. TrajectoryCrafter-type methods that rely on depth-based~\cite{hu2025depthcrafter} conditioning are prone to semantic drift and identity inconsistency. In the second column, due to insufficient conditioning coverage, TrajectoryCrafter\cite{yu2025trajectorycrafter} mistakenly interprets the reef region as part of the turtle. In contrast, ReCamMaster\cite{bai2025recammaster} injects camera poses directly as conditioning, without providing geometric constraints via target-view rerendering. As a result, it is more likely to produce hallucinations in regions with insufficient geometry. These observations suggest that in dynamic underwater novel view synthesis, relying solely on pose conditioning or incomplete geometric conditioning can only maintain cross-view and temporal consistency under small viewpoint changes. Our method mitigates these issues by improving geometric coverage and implicitly modeling the medium with the Medium-Aware Block.

\begin{figure*}[t]
	\centering
	\includegraphics[width=\linewidth]{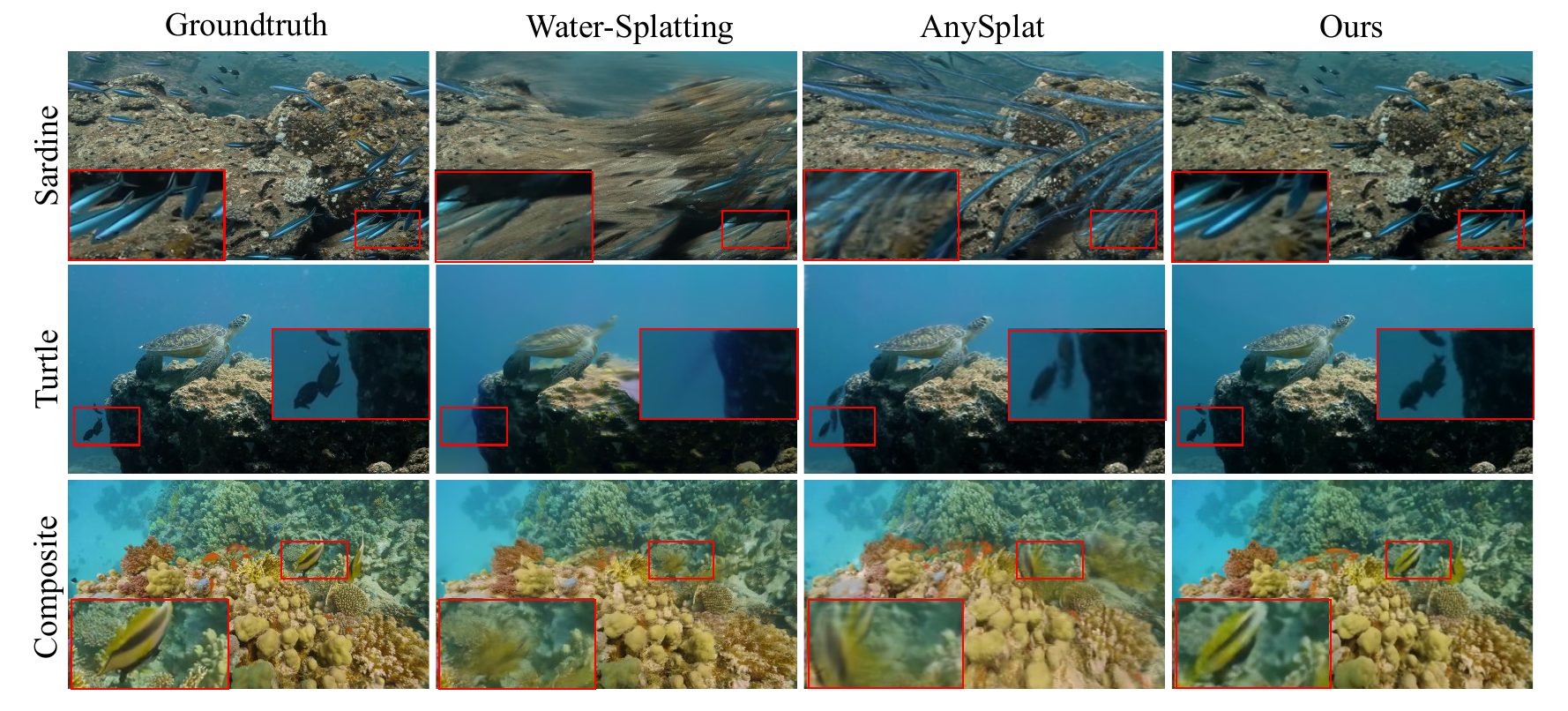}
	\caption{Qualitative comparison on NUSR. Red boxes highlight distractor regions with zoomed-in views. Our method achieves better reconstruction quality in both static regions and distractor regions.}
	\label{fig:nusr_qual}
    \vspace{-0.2cm}
\end{figure*}

\begin{table}[t]
	\centering
	\caption{Quantitative comparison on NUSR. PSNR, SSIM, and LPIPS comparisons on the four NUSR scenes. Best results are in \textbf{bold}, second best are \underline{underlined}.}
	\label{tab:nusr_quant}
	\scriptsize
	\setlength{\tabcolsep}{0.5pt}
	\begin{tabular}{lccc ccc ccc ccc}
		\toprule
		& \multicolumn{3}{c}{Sardine} & \multicolumn{3}{c}{Turtle} & \multicolumn{3}{c}{Composite} & \multicolumn{3}{c}{Coral} \\
		\cmidrule(lr){2-4}\cmidrule(lr){5-7}\cmidrule(lr){8-10}\cmidrule(lr){11-13}
		Method & PSNR$\uparrow$ & SSIM$\uparrow$ & LPIPS$\downarrow$
		& PSNR$\uparrow$ & SSIM$\uparrow$ & LPIPS$\downarrow$
		& PSNR$\uparrow$ & SSIM$\uparrow$ & LPIPS$\downarrow$
		& PSNR$\uparrow$ & SSIM$\uparrow$ & LPIPS$\downarrow$ \\
		\midrule
		3DGS~\cite{kerbl20233d} & 17.28 & 0.304 & 0.533 & 21.76 & 0.732 & 0.316 & 16.18 & 0.290 & 0.514 & 15.90 & 0.199 & 0.528 \\
		AnySplat~\cite{jiang2025anysplat} & 15.83 & 0.252 & 0.532 & 23.75 & 0.778 & 0.248 & 17.44 & 0.404 & 0.463 & 16.22 & 0.280 & 0.515 \\
		W.Splat.~\cite{li2025watersplatting} & \underline{18.14} & 0.351 & 0.476 & 22.74 & 0.774 & 0.331 & \textbf{19.55} & 0.468 & \textbf{0.237} & \underline{17.48} & 0.230 & \underline{0.316} \\
		Traj.C.~\cite{yu2025trajectorycrafter} & 17.08 & \underline{0.463} & \underline{0.313} & \underline{24.46} & \underline{0.823} & \underline{0.133} & 16.30 & \underline{0.563} & 0.401 & 16.33 & \underline{0.467} & 0.503 \\
		\textbf{Ours} & \textbf{19.58} & \textbf{0.517} & \textbf{0.248} & \textbf{28.36} & \textbf{0.853} & \textbf{0.104} & \underline{17.80} & \textbf{0.609} & \underline{0.340} & \textbf{18.76} & \textbf{0.554} & \textbf{0.254} \\
		\bottomrule
	\end{tabular}
    \vspace{-0.4cm}
\end{table}

\noindent\textbf{VBench Results.}
We further conduct quantitative evaluation on $18$ dynamic-view cases. As evaluation metrics, we adopt the VBench protocol~\cite{huang2024vbench}, including Subject Consistency, Background Consistency, Temporal Flickering, Motion Smoothness, Aesthetic Quality, and Imaging Quality. The results are shown in Tab.~\ref{tab:vbench_dynamic}. Overall, our method achieves better performance on consistency-related metrics, reaching $0.906$ in Subject Consis.\ and $0.920$ in Background Consis., outperforming both TrajectoryCrafter and ReCamMaster, indicating more stable spatiotemporal content and more reliable imaging quality under viewpoint changes.
We note that ReCamMaster achieves a higher Motion Smooth.\ score of $0.989$. This is because ReCamMaster uses direct pose injection and applies pose smoothing every four frames, which favors the motion smoothness metric. Considering that our setting requires strict view alignment and geometric consistency for static-scene, we do not adopt this strategy.

\begin{figure*}[t]
	\centering
	\includegraphics[width=\linewidth]{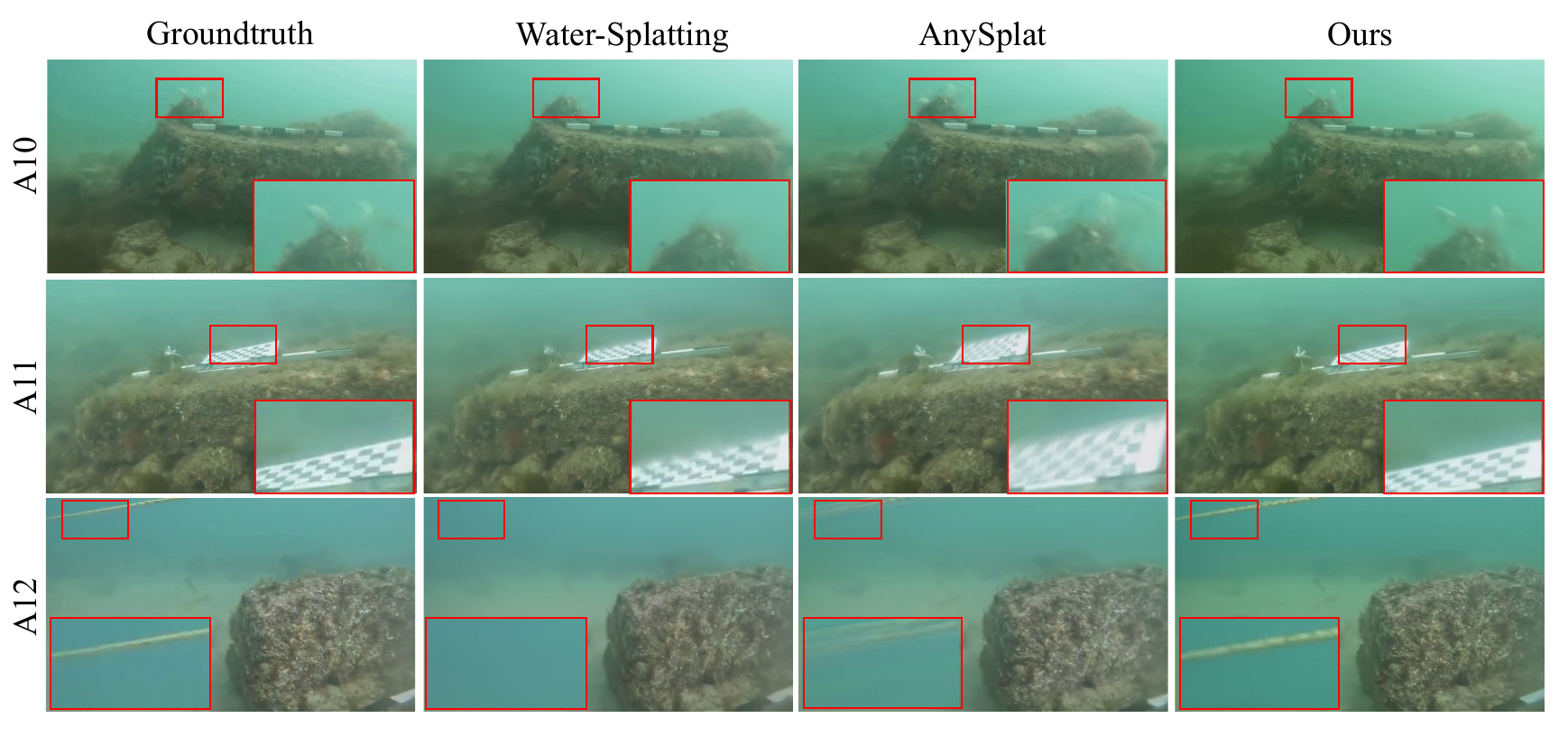}
	\caption{Qualitative comparison on DRUVA. Red boxes highlight detail regions that are strongly affected by water disturbances, including the flag in A11 and the rope in A12. }
	\label{fig:druva_qual}
    \vspace{-0.2cm}
\end{figure*}

\begin{table}[t]
	\centering
	\caption{Quantitative comparison on DRUVA. PSNR, SSIM, and LPIPS comparisons on four DRUVA scenes. Best results are in \textbf{bold}, second best are \underline{underlined}.}
	\label{tab:druva_quant}
	\scriptsize
	\setlength{\tabcolsep}{0.5pt}
	\begin{tabular}{lccc ccc ccc ccc}
		\toprule
		& \multicolumn{3}{c}{A10} & \multicolumn{3}{c}{A11} & \multicolumn{3}{c}{A12} & \multicolumn{3}{c}{A13} \\
		\cmidrule(lr){2-4}\cmidrule(lr){5-7}\cmidrule(lr){8-10}\cmidrule(lr){11-13}
		Method & PSNR$\uparrow$ & SSIM$\uparrow$ & LPIPS$\downarrow$
		& PSNR$\uparrow$ & SSIM$\uparrow$ & LPIPS$\downarrow$
		& PSNR$\uparrow$ & SSIM$\uparrow$ & LPIPS$\downarrow$
		& PSNR$\uparrow$ & SSIM$\uparrow$ & LPIPS$\downarrow$ \\
		\midrule
		3DGS~\cite{kerbl20233d} & 29.60 & \underline{0.914} & 0.297 & \underline{26.82} & 0.835 & 0.322 & \underline{26.59} & 0.861 & 0.303 & 23.91 & 0.853 & 0.368 \\
		AnySplat~\cite{jiang2025anysplat} & \textbf{31.56} & 0.920 & 0.268 & 26.19 & 0.846 & 0.320 & 25.97 & 0.861 & 0.302 & \textbf{27.52} & 0.894 & 0.346 \\
		W.Splat.~\cite{li2025watersplatting} & 30.03 & 0.933 & \textbf{0.086} & 25.10 & 0.856 & 0.196 & 25.38 & 0.882 & 0.209 & 25.13 & \textbf{0.947} & \textbf{0.119} \\
		Traj.C.~\cite{yu2025trajectorycrafter} & 22.55 & \underline{0.940} & \underline{0.129} & 21.48 & \underline{0.900} & \underline{0.174} & 19.79 & \underline{0.894} & \underline{0.193} & 19.40 & 0.919 & 0.202 \\
		ReCam.~\cite{bai2025recammaster} & 24.23 & 0.906 & 0.269 & 22.98 & 0.847 & 0.296 & 23.87 & 0.880 & 0.259 & 23.31 & 0.895 & 0.273 \\
		\textbf{Ours} & \underline{30.23} & \textbf{0.955} & 0.132 & \textbf{27.86} & \textbf{0.913} & \textbf{0.123} & \textbf{26.69} & \textbf{0.921} & \textbf{0.140} & \underline{27.00} & \underline{0.941} & \underline{0.133} \\
		\bottomrule
	\end{tabular}
    \vspace{-0.4cm}
\end{table}

\subsection{Evaluation on Static Scenes}
\label{sec:exp_static}


We also evaluate our method for static scenes on NUSR \cite{tang2024neural} and DRUVA \cite{varghese2023self}. The corresponding qualitative and quantitative results are as follows.

\noindent\textbf{Evaluation on NUSR.}
Fig.~\ref{fig:nusr_qual} shows qualitative comparisons with Water-Splatting \cite{li2025watersplatting} and AnySplat \cite{jiang2025anysplat}. Although existing methods can generally recover background structures well, they typically struggle with small, dynamic distractors that frequently appear in underwater videos,\textit{ e.g.,} small fish. In the \textit{Turtle} scene, the subtle motion of a small fish in the lower-left corner poses a challenge: AnySplat tends to introduce artifacts, while WaterSplatting often collapses the region into an averaged color patch, causing the loss of fine details. Similar behavior is observed for the tropical fish in the \textit{Composite} and \textit{Sardine} scenes. In the zoomed-in views, both WaterSplatting and AnySplat are more likely to fail under severe distractors. In contrast, our method not only reconstructs the seabed background and main structures clearly, but also adapts better to such distractors, yielding more stable reconstructions.

Tab.~\ref{tab:nusr_quant} reports quantitative results on the four NUSR scenes~\cite{tang2024neural}. Overall, our method achieves the best performance on all scenes except \textit{Composite}. In the \textit{Turtle} scene, our method improves PSNR to $28.36$ and SSIM to $0.853$, while reducing LPIPS to $0.104$, indicating that it maintains more stable reconstruction quality even in the presence of small dynamic distractors.

\begin{figure*}[t]
	\centering
	\includegraphics[width=0.95\linewidth]{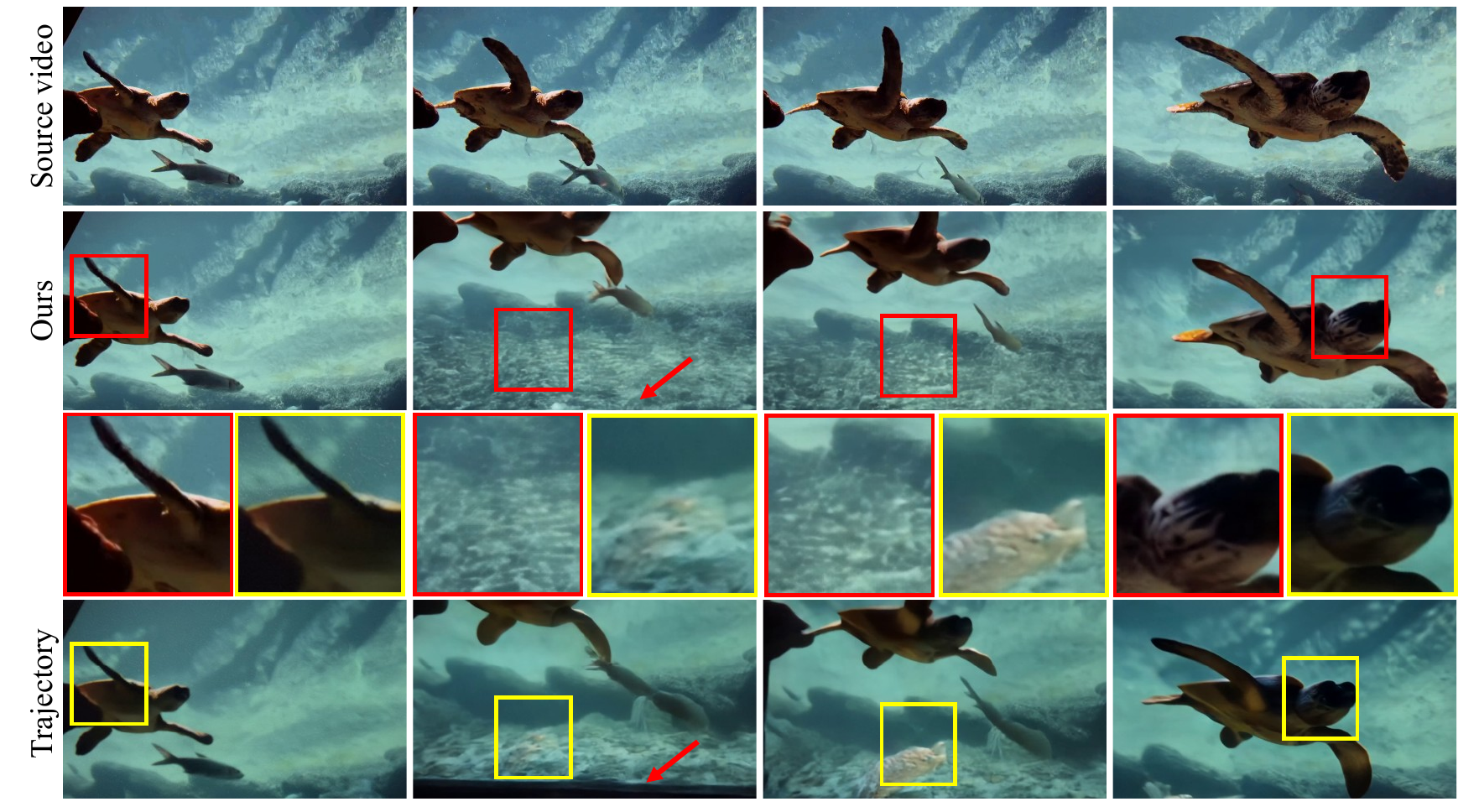}
	\caption{\textbf{Hallucination analysis on dynamic scenes.} Red boxes and arrows highlight failure cases under large viewpoint shifts and unseen regions.}
	\label{fig:dynamic_detail}
    \vspace{-0.4cm}
\end{figure*}

\noindent\textbf{Evaluation on DRUVA.}
Fig.~\ref{fig:druva_qual} shows qualitative comparisons on DRUVA~\cite{varghese2023self}. In the A11 scene, the red box highlights the calibration flag region, where WaterSplatting\cite{li2025watersplatting} and AnySplat\cite{jiang2025anysplat} exhibit varying degrees of trailing artifacts and blur, weakening edge details and texture structures. In the A12 scene, the red box corresponds to a thin rope structure. WaterSplatting fails to reconstruct this detail, while AnySplat produces visible breaks and discontinuities. In contrast, our method preserves fine structures and generates more consistent geometry and appearance, under water disturbances and cross-view medium variation.

Tab.~\ref{tab:druva_quant} reports quantitative results on four DRUVA scenes. Overall, our method achieves the best performance on most metrics. In the A11 scene, our method improves PSNR to $27.861$ and SSIM to $0.913$, while reducing LPIPS to $0.123$, indicating more stable reconstruction quality.

\subsection{Ablation Studies}
\label{sec:exp_ablation}

\noindent\textbf{Dynamic Scene Analysis.}
Fig.~\ref{fig:dynamic_detail} provides a closer look at how Ocean4D benefits from 4D-GCC and the Medium-Aware Block. 
In the first and fourth columns, TrajectoryCrafter~\cite{yu2025trajectorycrafter} synthesizes a turtle that is overall dim and fails to preserve the local highlights caused by non-uniform underwater illumination. Our results better match the lighting pattern in the source video and retain sharper texture details. This is consistent with the role of the Medium-Aware Block, which injects depth-aware modulation into the denoising pathway to better handle absorption- and scattering-induced appearance changes and suppress color drifting.
The second and third columns emphasize the need for complete geometric conditioning under large viewpoint shifts. 
In the second column, TrajectoryCrafter shows missing content in the red-arrow region, whereas our method completes it with a consistent seabed structure. In the third column, TrajectoryCrafter produces fish-like hallucinations on the seabed. By contrast, Ocean4D leverages 4D-GCC to aggregate complementary cross-frame geometric evidence, which improves conditioning coverage and reduces hallucinations.

\begin{figure*}[t]
	\centering
	\includegraphics[width=\linewidth]{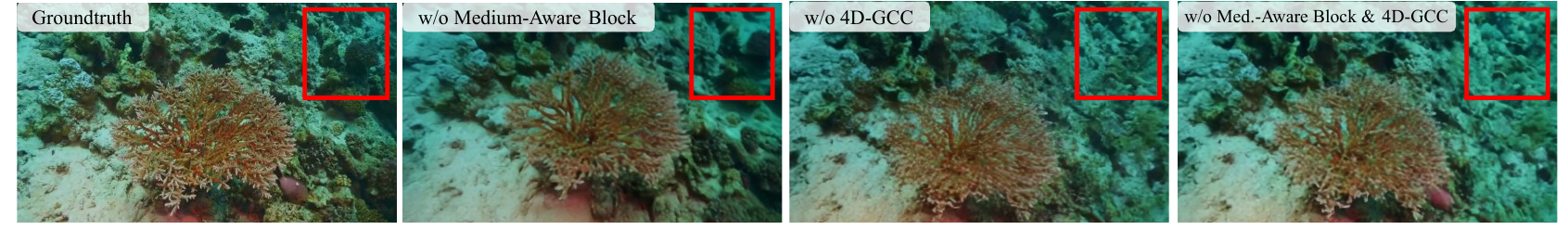}
	\caption{\textbf{Ablation visualization.} Visual comparisons on static scenes, including Groundtruth, w/o Medium-Aware Block, w/o 4D-GCC, and w/o Medium-Aware Block \& 4D-GCC. Red boxes highlight differences in reef textures and fine details.}
	\label{fig:ablation_qual}
    \vspace{-0.2cm}
\end{figure*}

\begin{table}[t]
	\centering
    \scriptsize
	\caption{\textbf{Ablation results.} Average metrics on the NUSR static scenes.}
	\label{tab:ablation_quant}
	\setlength{\tabcolsep}{6pt}
	\begin{tabular}{lccc}
		\toprule
		Method & PSNR$\uparrow$ & SSIM$\uparrow$ & LPIPS$\downarrow$ \\
		\midrule
		Ours w/o 4D-GCC & 19.782 & 0.573 & 0.295 \\
		Ours w/o Medium-Aware Block & 18.710 & 0.557 & 0.306 \\
		Ours w/o 4D-GCC \& Medium-Aware Block & 18.543 & 0.579 & 0.337 \\
		Ours & \textbf{21.125} & \textbf{0.633} & \textbf{0.237} \\
		\bottomrule
	\end{tabular}
    \vspace{-0.4cm}
\end{table}

\noindent\textbf{Static Scene Analysis.}
We also conduct ablation studies on static scenes to evaluate the contributions of 4D-GCC and the Medium-Aware Block. 
Fig.~\ref{fig:ablation_qual} presents representative visual comparisons. With 4D-GCC, reef and seabed structures are reconstructed more completely, indicating improved cross-view geometric consistency from more reliable conditioning coverage. With the Medium-Aware Block, underwater appearance becomes more stable. Although geometric consistency degrades without the assistance of 4D-GCC and the reef structure becomes incomplete, the modeled details are sharper, with stronger color consistency , suggesting that the Medium-Aware Block helps absorb medium variation during diffusion denoising and suppress appearance drift induced by scattering and attenuation. Tab.~\ref{tab:ablation_quant} reports quantitative results. The full model achieves the best performance on PSNR, SSIM, and LPIPS, reaching $21.125$, $0.633$, and $0.237$, respectively, further validating the effectiveness of our design.

\section{Conclusion}
\label{sec:conclusion}

We propose \textbf{Ocean4D}, a unified framework for \textbf{Generative 4D Reconstruction} from monocular underwater videos. Ocean4D leverages \textbf{4D-GCC} to construct 4D geometrically consistent conditioning with improved cross-frame coverage under occlusions and large viewpoint changes. We further introduce a \textbf{Medium-Aware Block} to inject depth-aware modulation into the denoising process, stabilizing underwater appearance variations caused by absorption and scattering. Experiments on both dynamic and static underwater scenes demonstrate that Ocean4D achieves state-of-the-art performance in terms of geometric consistency and underwater visual quality, offering a practical and effective solution for real-world underwater 4D modeling.


%
%
\bibliographystyle{splncs04}
\bibliography{main}

\begin{thebibliography}{10}
\providecommand{\url}[1]{\texttt{#1}}
\providecommand{\urlprefix}{URL }
\providecommand{\doi}[1]{https://doi.org/#1}

\bibitem{akkaynak2018revised}
Akkaynak, D., Treibitz, T.: A revised underwater image formation model. In: Proc. CVPR. pp. 6723--6732 (2018)

\bibitem{akkaynak2019sea}
Akkaynak, D., Treibitz, T.: Sea-thru: A method for removing water from underwater images. In: Proc. CVPR. pp. 1682--1691 (2019)

\bibitem{arain2020close}
Arain, B., Dayoub, F., Rigby, P., Dunbabin, M.: Close-proximity underwater terrain mapping using learning-based coarse range estimation. arXiv preprint arXiv:2001.00330  (2020)

\bibitem{bai2025recammaster}
Bai, J., Xia, M., Fu, X., Wang, X., Mu, L., Cao, J., Liu, Z., Hu, H., Bai, X., Wan, P., et~al.: Recammaster: Camera-controlled generative rendering from a single video. In: Proc. ICCV. pp. 14834--14844 (2025)

\bibitem{barron2021mip}
Barron, J.T., Mildenhall, B., Tancik, M., Hedman, P., Martin-Brualla, R., Srinivasan, P.P.: Mip-nerf: A multiscale representation for anti-aliasing neural radiance fields. In: Proc. ICCV. pp. 5855--5864 (2021)

\bibitem{barron2022mip}
Barron, J.T., Mildenhall, B., Verbin, D., Srinivasan, P.P., Hedman, P.: Mip-nerf 360: Unbounded anti-aliased neural radiance fields. In: Proc. CVPR. pp. 5470--5479 (2022)

\bibitem{barron2023zip}
Barron, J.T., Mildenhall, B., Verbin, D., Srinivasan, P.P., Hedman, P.: Zip-nerf: Anti-aliased grid-based neural radiance fields. In: Proc. ICCV. pp. 19697--19705 (2023)

\bibitem{blattmann2023stable}
Blattmann, A., Dockhorn, T., Kulal, S., Mendelevitch, D., Kilian, M., Lorenz, D., Levi, Y., English, Z., Voleti, V., Letts, A., et~al.: Stable video diffusion: Scaling latent video diffusion models to large datasets. arXiv preprint arXiv:2311.15127  (2023)

\bibitem{blattmann2023align}
Blattmann, A., Rombach, R., Ling, H., Dockhorn, T., Kim, S.W., Fidler, S., Kreis, K.: Align your latents: High-resolution video synthesis with latent diffusion models. In: Proc. CVPR. pp. 22563--22575 (2023)

\bibitem{chen2021mvsnerf}
Chen, A., Xu, Z., Zhao, F., Zhang, X., Xiang, F., Yu, J., Su, H.: Mvsnerf: Fast generalizable radiance field reconstruction from multi-view stereo. In: Proc. ICCV. pp. 14124--14133 (2021)

\bibitem{chen2023videocrafter1}
Chen, H., Xia, M., He, Y., Zhang, Y., Cun, X., Yang, S., Xing, J., Liu, Y., Chen, Q., Wang, X., et~al.: Videocrafter1: Open diffusion models for high-quality video generation. arXiv preprint arXiv:2310.19512  (2023)

\bibitem{chen2024sp}
Chen, L., Xiong, Y., Zhang, Y., Yu, R., Fang, L., Liu, D.: Sp-seanerf: Underwater neural radiance fields with strong scattering perception. Computers \& Graphics  \textbf{123},  104025 (2024)

\bibitem{chen2024mvsplat}
Chen, Y., Xu, H., Zheng, C., Zhuang, B., Pollefeys, M., Geiger, A., Cham, T.J., Cai, J.: Mvsplat: Efficient 3d gaussian splatting from sparse multi-view images. In: Proc. ECCV. pp. 370--386. Springer (2024)

\bibitem{du2024udr}
Du, Y., Zhang, Z., Zhang, P., Sun, F., Lv, X.: Udr-gs: Enhancing underwater dynamic scene reconstruction with depth regularization. Symmetry  \textbf{16}(8), ~1010 (2024)

\bibitem{duan20244d}
Duan, Y., Wei, F., Dai, Q., He, Y., Chen, W., Chen, B.: 4d-rotor gaussian splatting: towards efficient novel view synthesis for dynamic scenes. In: proc. ACM SIGGRAPH. pp. 1--11 (2024)

\bibitem{fan2024instantsplat}
Fan, Z., Cong, W., Wen, K., Wang, K., Zhang, J., Ding, X., Xu, D., Ivanovic, B., Pavone, M., Pavlakos, G., et~al.: Instantsplat: Sparse-view gaussian splatting in seconds. arXiv preprint arXiv:2403.20309  (2024)

\bibitem{feng2025ae}
Feng, C., Yu, W., Cheng, X., Tang, Z., Zhang, J., Yuan, L., Tian, Y.: Ae-nerf: Augmenting event-based neural radiance fields for non-ideal conditions and larger scenes. In: Proc. AAAI. vol.~39, pp. 2924--2932 (2025)

\bibitem{fu2014retinex}
Fu, X., Zhuang, P., Huang, Y., Liao, Y., Zhang, X.P., Ding, X.: A retinex-based enhancing approach for single underwater image. In: Proc. ICIP. pp. 4572--4576. IEEE (2014)

\bibitem{gao2024cat3d}
Gao*, R., Holynski*, A., Henzler, P., Brussee, A., Martin-Brualla, R., Srinivasan, P.P., Barron, J.T., Poole*, B.: Cat3d: Create anything in 3d with multi-view diffusion models. Advances in Neural Information Processing Systems  (2024)

\bibitem{ghani2014underwater}
Ghani, A.S.A., Isa, N.A.M.: Underwater image quality enhancement through rayleigh-stretching and averaging image planes. International Journal of Naval Architecture and Ocean Engineering  \textbf{6}(4),  840--866 (2014)

\bibitem{hernandez2016autonomous}
Hern{\'a}ndez, J.D., Isteni{\v{c}}, K., Gracias, N., Palomeras, N., Campos, R., Vidal, E., Garcia, R., Carreras, M.: Autonomous underwater navigation and optical mapping in unknown natural environments. Sensors  \textbf{16}(8), ~1174 (2016)

\bibitem{hopkinson2020automated}
Hopkinson, B.M., King, A.C., Owen, D.P., Johnson-Roberson, M., Long, M.H., Bhandarkar, S.M.: Automated classification of three-dimensional reconstructions of coral reefs using convolutional neural networks. PloS one  \textbf{15}(3),  e0230671 (2020)

\bibitem{hu2023overview}
Hu, K., Wang, T., Shen, C., Weng, C., Zhou, F., Xia, M., Weng, L.: Overview of underwater 3d reconstruction technology based on optical images. Journal of Marine Science and Engineering  \textbf{11}(5), ~949 (2023)

\bibitem{hu2025depthcrafter}
Hu, W., Gao, X., Li, X., Zhao, S., Cun, X., Zhang, Y., Quan, L., Shan, Y.: Depthcrafter: Generating consistent long depth sequences for open-world videos. In: Proc. CVPR. pp. 2005--2015 (2025)

\bibitem{huang2024vbench}
Huang, Z., He, Y., Yu, J., Zhang, F., Si, C., Jiang, Y., Zhang, Y., Wu, T., Jin, Q., Chanpaisit, N., et~al.: Vbench: Comprehensive benchmark suite for video generative models. In: Proc. CVPR. pp. 21807--21818 (2024)

\bibitem{iqbal2010enhancing}
Iqbal, K., Odetayo, M., James, A., Salam, R.A., Talib, A.Z.H.: Enhancing the low quality images using unsupervised colour correction method. In: 2010 IEEE international conference on systems, man and cybernetics. pp. 1703--1709. IEEE (2010)

\bibitem{jiang2025anysplat}
Jiang, L., Mao, Y., Xu, L., Lu, T., Ren, K., Jin, Y., Xu, X., Yu, M., Pang, J., Zhao, F., et~al.: Anysplat: Feed-forward 3d gaussian splatting from unconstrained views. ACM Transactions on Graphics  \textbf{44}(6),  1--16 (2025)

\bibitem{kerbl20233d}
Kerbl, B., Kopanas, G., Leimk{\"u}hler, T., Drettakis, G., et~al.: 3d gaussian splatting for real-time radiance field rendering. ACM Trans. Graph.  \textbf{42}(4),  139--1 (2023)

\bibitem{levy2023seathru}
Levy, D., Peleg, A., Pearl, N., Rosenbaum, D., Akkaynak, D., Korman, S., Treibitz, T.: Seathru-nerf: Neural radiance fields in scattering media. In: Proc. CVPR. pp. 56--65 (2023)

\bibitem{li2025watersplatting}
Li, H., Song, W., Xu, T., Elsig, A., Kulhanek, J.: Watersplatting: Fast underwater 3d scene reconstruction using gaussian splatting. In: Proc. 3DV. pp. 969--978. IEEE (2025)

\bibitem{li2022blip}
Li, J., Li, D., Xiong, C., Hoi, S.: Blip: Bootstrapping language-image pre-training for unified vision-language understanding and generation. In: Proc. ICML. pp. 12888--12900. PMLR (2022)

\bibitem{liang2024analytic}
Liang, Z., Zhang, Q., Hu, W., Zhu, L., Feng, Y., Jia, K.: Analytic-splatting: Anti-aliased 3d gaussian splatting via analytic integration. In: Proc. ECCV. pp. 281--297. Springer (2024)

\bibitem{lin2024open}
Lin, B., Ge, Y., Cheng, X., Li, Z., Zhu, B., Wang, S., He, X., Ye, Y., Yuan, S., Chen, L., et~al.: Open-sora plan: Open-source large video generation model. arXiv preprint arXiv:2412.00131  (2024)

\bibitem{lin2021barf}
Lin, C.H., Ma, W.C., Torralba, A., Lucey, S.: Barf: Bundle-adjusting neural radiance fields. In: Proc. ICCV. pp. 5741--5751 (2021)

\bibitem{liu2023zero}
Liu, R., Wu, R., Van~Hoorick, B., Tokmakov, P., Zakharov, S., Vondrick, C.: Zero-1-to-3: Zero-shot one image to 3d object. In: Proc. ICCV. pp. 9298--9309 (2023)

\bibitem{liu2024aquatic}
Liu, S., Lu, J., Gu, Z., Li, J., Deng, Y.: Aquatic-gs: A hybrid 3d representation for underwater scenes. arXiv preprint arXiv:2411.00239  (2024)

\bibitem{mildenhall2021nerf}
Mildenhall, B., Srinivasan, P.P., Tancik, M., Barron, J.T., Ramamoorthi, R., Ng, R.: Nerf: Representing scenes as neural radiance fields for view synthesis. Communications of the ACM  \textbf{65}(1),  99--106 (2021)

\bibitem{muller2024multidiff}
M{\"u}ller, N., Schwarz, K., R{\"o}ssle, B., Porzi, L., Bulo, S.R., Nie{\ss}ner, M., Kontschieder, P.: Multidiff: Consistent novel view synthesis from a single image. In: Proc. CVPR. pp. 10258--10268 (2024)

\bibitem{de2022high}
de~Oliveira, L.M.C., Lim, A., Conti, L.A., Wheeler, A.J.: High-resolution 3d mapping of cold-water coral reefs using machine learning. Frontiers in Environmental Science  \textbf{10},  1044706 (2022)

\bibitem{park2021nerfies}
Park, K., Sinha, U., Barron, J.T., Bouaziz, S., Goldman, D.B., Seitz, S.M., Martin-Brualla, R.: Nerfies: Deformable neural radiance fields. In: Proc. ICCV. pp. 5865--5874 (2021)

\bibitem{pumarola2021d}
Pumarola, A., Corona, E., Pons-Moll, G., Moreno-Noguer, F.: D-nerf: Neural radiance fields for dynamic scenes. In: Proc. CVPR. pp. 10318--10327 (2021)

\bibitem{raffel2020exploring}
Raffel, C., Shazeer, N., Roberts, A., Lee, K., Narang, S., Matena, M., Zhou, Y., Li, W., Liu, P.J.: Exploring the limits of transfer learning with a unified text-to-text transformer. Journal of Machine Learning Research  \textbf{21}(140),  1--67 (2020)

\bibitem{ren2025gen3c}
Ren, X., Shen, T., Huang, J., Ling, H., Lu, Y., Nimier-David, M., M{\"u}ller, T., Keller, A., Fidler, S., Gao, J.: Gen3c: 3d-informed world-consistent video generation with precise camera control. In: Proc. CVPR. pp. 6121--6132 (2025)

\bibitem{sargent2024zeronvs}
Sargent, K., Li, Z., Shah, T., Herrmann, C., Yu, H.X., Zhang, Y., Chan, E.R., Lagun, D., Fei-Fei, L., Sun, D., et~al.: Zeronvs: Zero-shot 360-degree view synthesis from a single image. In: Proc. CVPR. pp. 9420--9429 (2024)

\bibitem{sethuraman2023waternerf}
Sethuraman, A.V., Ramanagopal, M.S., Skinner, K.A.: Waternerf: Neural radiance fields for underwater scenes. In: OCEANS 2023-MTS/IEEE US Gulf Coast. pp.~1--7. IEEE (2023)

\bibitem{song2020denoising}
Song, J., Meng, C., Ermon, S.: Denoising diffusion implicit models. In: Proc. ICLR (2020)

\bibitem{song2023nerfplayer}
Song, L., Chen, A., Li, Z., Chen, Z., Chen, L., Yuan, J., Xu, Y., Geiger, A.: Nerfplayer: A streamable dynamic scene representation with decomposed neural radiance fields. IEEE Transactions on Visualization and Computer Graphics  \textbf{29}(5),  2732--2742 (2023)

\bibitem{sucar2026v}
Sucar, E., Insafutdinov, E., Lai, Z., Vedaldi, A.: V-dpm: 4d video reconstruction with dynamic point maps. arXiv preprint arXiv:2601.09499  (2026)

\bibitem{tang2024neural}
Tang, Y., Zhu, C., Wan, R., Xu, C., Shi, B.: Neural underwater scene representation. In: Proc. CVPR. pp. 11780--11789 (2024)

\bibitem{van2024generative}
Van~Hoorick, B., Wu, R., Ozguroglu, E., Sargent, K., Liu, R., Tokmakov, P., Dave, A., Zheng, C., Vondrick, C.: Generative camera dolly: Extreme monocular dynamic novel view synthesis. In: Proc. ECCV. pp. 313--331. Springer (2024)

\bibitem{varghese2023self}
Varghese, N., Kumar, A., Rajagopalan, A.: Self-supervised monocular underwater depth recovery, image restoration, and a real-sea video dataset. In: Proc. ICCV. pp. 12248--12258 (2023)

\bibitem{verbin2024ref}
Verbin, D., Hedman, P., Mildenhall, B., Zickler, T., Barron, J.T., Srinivasan, P.P.: Ref-nerf: Structured view-dependent appearance for neural radiance fields. IEEE Transactions on Pattern Analysis and Machine Intelligence  \textbf{47}(11),  9426--9437 (2024)

\bibitem{wang20253d}
Wang, H., Agapito, L.: 3d reconstruction with spatial memory. In: Proc. 3DV. pp. 78--89. IEEE (2025)

\bibitem{wang2025vggt}
Wang, J., Chen, M., Karaev, N., Vedaldi, A., Rupprecht, C., Novotny, D.: Vggt: Visual geometry grounded transformer. In: Proc. CVPR. pp. 5294--5306 (2025)

\bibitem{wang2025continuous}
Wang, Q., Zhang, Y., Holynski, A., Efros, A.A., Kanazawa, A.: Continuous 3d perception model with persistent state. In: Proc. CVPR. pp. 10510--10522 (2025)

\bibitem{wang2024dust3r}
Wang, S., Leroy, V., Cabon, Y., Chidlovskii, B., Revaud, J.: Dust3r: Geometric 3d vision made easy. In: Proc. CVPR. pp. 20697--20709 (2024)

\bibitem{wu20244d}
Wu, G., Yi, T., Fang, J., Xie, L., Zhang, X., Wei, W., Liu, W., Tian, Q., Wang, X.: 4d gaussian splatting for real-time dynamic scene rendering. In: Proc. CVPR. pp. 20310--20320 (2024)

\bibitem{wu2025cat4d}
Wu, R., Gao, R., Poole, B., Trevithick, A., Zheng, C., Barron, J.T., Holynski, A.: Cat4d: Create anything in 4d with multi-view video diffusion models. In: Proc. CVPR. pp. 26057--26068 (2025)

\bibitem{wu2024reconfusion}
Wu, R., Mildenhall, B., Henzler, P., Park, K., Gao, R., Watson, D., Srinivasan, P.P., Verbin, D., Barron, J.T., Poole, B., et~al.: Reconfusion: 3d reconstruction with diffusion priors. In: Proc. CVPR. pp. 21551--21561 (2024)

\bibitem{xie2024uveb}
Xie, Y., Kong, L., Chen, K., Zheng, Z., Yu, X., Yu, Z., Zheng, B.: Uveb: A large-scale benchmark and baseline towards real-world underwater video enhancement. In: Proc. CVPR. pp. 22358--22367 (2024)

\bibitem{yang2025seasplat}
Yang, D., Leonard, J.J., Girdhar, Y.: Seasplat: Representing underwater scenes with 3d gaussian splatting and a physically grounded image formation model. In: Proc. ICRA. pp. 7632--7638. IEEE (2025)

\bibitem{yang2024cogvideox}
Yang, Z., Teng, J., Zheng, W., Ding, M., Huang, S., Xu, J., Yang, Y., Hong, W., Zhang, X., Feng, G., et~al.: Cogvideox: Text-to-video diffusion models with an expert transformer. In: Proc. ICLR (2024)

\bibitem{yariv2021volume}
Yariv, L., Gu, J., Kasten, Y., Lipman, Y.: Volume rendering of neural implicit surfaces. Advances in Neural Information Processing Systems  \textbf{34},  4805--4815 (2021)

\bibitem{yu2025trajectorycrafter}
Yu, M., Hu, W., Xing, J., Shan, Y.: Trajectorycrafter: Redirecting camera trajectory for monocular videos via diffusion models. In: Proc. ICCV. pp. 100--111 (2025)

\bibitem{yu2023nofa}
Yu, W., Fan, Y., Zhang, Y., Wang, X., Yin, F., Bai, Y., Cao, Y.P., Shan, Y., Wu, Y., Sun, Z., et~al.: Nofa: Nerf-based one-shot facial avatar reconstruction. In: Proc. ACM SIGGRAPH. pp. 1--12 (2023)

\bibitem{yu2024viewcrafter}
Yu, W., Xing, J., Yuan, L., Hu, W., Li, X., Huang, Z., Gao, X., Wong, T.T., Shan, Y., Tian, Y.: Viewcrafter: Taming video diffusion models for high-fidelity novel view synthesis. IEEE Transactions on Pattern Analysis and Machine Intelligence  (2025)

\bibitem{zhang2024monst3r}
Zhang, J., Herrmann, C., Hur, J., Jampani, V., Darrell, T., Cole, F., Sun, D., Yang, M.H.: Monst3r: A simple approach for estimating geometry in the presence of motion. In: Proc. ICLR (2024)

\bibitem{zhang2017underwater}
Zhang, S., Wang, T., Dong, J., Yu, H.: Underwater image enhancement via extended multi-scale retinex. Neurocomputing  \textbf{245}, ~1--9 (2017)

\bibitem{zhu2024fsgs}
Zhu, Z., Fan, Z., Jiang, Y., Wang, Z.: Fsgs: Real-time few-shot view synthesis using gaussian splatting. In: Proc. ECCV. pp. 145--163. Springer (2024)

\end{thebibliography}

\clearpage

\end{document}